\documentclass[sigconf]{acmart}

\copyrightyear{2021}
\acmYear{2021}
\setcopyright{iw3c2w3}
\acmConference[WWW '21]{Proceedings of the Web Conference 2021}{April 19--23, 2021}{Ljubljana, Slovenia}
\acmBooktitle{Proceedings of the Web Conference 2021 (WWW '21), April 19--23, 2021, Ljubljana, Slovenia}

\acmPrice{}
\acmDOI{10.1145/3442381.3449889}
\acmISBN{978-1-4503-8312-7/21/04}

\AtBeginDocument{%
  \providecommand\BibTeX{{%
    \normalfont B\kern-0.5em{\scshape i\kern-0.25em b}\kern-0.8em\TeX}}}

\usepackage{makecell}
\usepackage{amsmath}
\usepackage[normalem]{ulem}

\DeclareMathOperator{\iC}{C}

\DeclareMathOperator{\iR}{R}

\newtheorem{assumption}{Assumption}
\usepackage{tcolorbox}
\usepackage{caption}
\usepackage{subcaption}
\usepackage{enumerate}
\usepackage[ruled,vlined]{algorithm2e}

\newcommand{\mmdel}[1]{}

\begin{document}

\title{Towards Content Provider Aware Recommender Systems}
\subtitle{A Simulation Study on the Interplay between User and Provider Utilities}

\author{Ruohan Zhan$^*$, Konstantina Christakopoulou$^\dagger$, Ya Le$^\dagger$, Jayden Ooi$^\dagger$, Martin Mladenov$^\dagger$, Alex Beutel$^\dagger$, Craig Boutilier$^\dagger$, Ed H. Chi$^\dagger$, Minmin Chen$^\dagger$}
\affiliation{
\institution{rhzhan@stanford.edu; $\{$konchris, elainele, jayden, mmladenov, alexbeutel,  cboutilier, edchi, minminc$\}$@google.com}
\country{}
}
\affiliation{%
  \institution{$^*$Stanford University, Stanford, CA}
  \country{}
}
\affiliation{%
  \institution{$^\dagger$Google, Mountain View, CA}
 \country{}
}

\renewcommand{\shortauthors}{Zhan et al.}

\begin{abstract}
Most existing recommender systems focus primarily on matching users (content consumers) to content which maximizes user satisfaction on the platform. It is increasingly obvious, however, that content providers have a critical  influence on user satisfaction through content creation, largely determining the content pool available for recommendation. A natural question thus arises: can we design recommenders taking into account the long-term utility of both users and content providers? By doing so, we hope to sustain more content providers and a more diverse content pool for long-term user satisfaction. 
Understanding the full impact of recommendations on both user and content provider groups is challenging.  This paper aims to serve as a research investigation of one approach toward building a content provider aware recommender, and evaluating its impact in a simulated setup. 

To characterize the user-recommender-provider interdependence, we complement user modeling by formalizing provider dynamics as well. The resulting joint dynamical system gives rise to  a  weakly-coupled partially observable Markov decision process driven by recommender actions and user feedback to providers. We then build a REINFORCE recommender agent, coined \texttt{EcoAgent}, to optimize a joint objective of user utility and the counterfactual utility lift of the content provider associated with the recommended content, which we show to be equivalent to maximizing overall user utility and the utilities of all content providers on the platform under some mild assumptions. To evaluate our approach, we  introduce a simulation environment capturing the key interactions among users, providers, and the recommender. We offer a number of simulated experiments that shed light on both the benefits \emph{and the limitations} of our approach. These results help understand \emph{how} and \emph{when} a content provider aware recommender agent is of benefit in building multi-stakeholder recommender systems.

\end{abstract}




\maketitle

\section{Introduction}
\label{sec:intro}
Recommender systems have been playing an increasingly important role for online services by matching users with content for personalization and experience enhancement. Most recommender systems exclusively focus on  serving content that maximizes user (consumer) satisfaction. 
Aligning a recommender system's objective with user utility is most natural given that users are first-hand consumers of recommendation services. However, another key group of players in any platform are the \emph{content providers}, who are also largely influenced by the recommendation decisions. 
In a purely user-centric design, the exposure given to different content providers can be vastly disproportionate, with a small subpopulation receiving the majority of attention \cite{mehrotra2018towards}.
When this happens, less established content providers  
can find it difficult to break the popularity barrier and may eventually leave the platform due to lack of exposure \cite{mladenov2020optimizing}. This is commonly referred to in marketplaces as the `superstar economy' \cite{rosen1981economics}, 
which in the long run can hurt the overall user utility 
by limiting the diversity of the pool of recommendable content.

As a result, there has been growing research interest in moving from a user-centric recommendation paradigm toward a multi-stakeholder setup \cite{burke2016towards, abdollahpouri2017recommender}, under which the viability of content providers on the platform, as a function of their respective satisfaction, is also considered.
Such content provider awareness can be achieved by maximizing user satisfaction with the constraint of keeping existing content providers on the platform \citep{mladenov2020optimizing}, or by trading off content relevance with content provider fairness~\citep{mehrotra2018towards}. However, most existing work treats content providers as static entities. Here, we consider the dynamics of content providers in response to recommendation, and propose a reinforcement learning (RL)-based recommendation approach to optimize the combination of both user and provider satisfaction. 

\begin{table*}[!tb]
  \caption{Related Work Comparison}
  \label{tab:related_work}
  \begin{tabular}{cllll}
    \toprule
    & \makecell{User \\ Modeling} & \makecell{Content Provider\\ Modeling} & \makecell{Counterfactual Reasoning} & \makecell{Joint User \& \\ Provider Optimization } \\
    \midrule
    Advertising Auctions \citep{vickrey1961counterspeculation,edelman2007internet,athey2010structural} &No & Yes & Yes & No \\
    Provider Fairness in  RecSys  \citep{biega2018equity,singh2018fairness,beutel2019fairness,mladenov2020optimizing,mehrotra2018towards} & Yes & No & Yes (in \cite{mehrotra2018towards}) & No \\
    \texttt{EcoAgent} (This paper) &  Yes &Yes & Yes & Yes\\
  \bottomrule
\end{tabular}
\end{table*}

Optimizing a recommendation policy for content provider satisfaction is significantly more challenging than for user satisfaction for two main reasons: (i) lack of a clear definition of \emph{provider utility}; and (ii) scalability. 
As opposed to more well-studied marketplaces, such as advertising platforms where suppliers clearly define their bidding policies, thus leaving signals indicating their utility \cite{athey2010structural}, content provider utility on recommendation platforms has received little attention. In this paper, we consider one such definition: we assume that provider utilities can be manifested through their actions e.g., via their content uploading frequencies, their decision to stay/leave the platform, or more generally how they respond to recommender actions and user feedback.\footnote{Importantly, our experiments highlight the importance of investigating different provider utility definitions under different recommendation setups.} 
This stands in contrast to existing work in fairness of recommender systems (e.g. \cite{singh2018fairness} and references therein), where the direct influence of platforms/users on provider utility is not considered.
Furthermore, in contrast to traditional two-sided marketplaces, such as \citep{athey2010structural}, where advertiser bidding policies are more or less static, 
content providers in recommender systems tend to be more adaptive to the environment. 
For example, based on the popularity of their existing content, providers may adapt their next content focus. To capture such considerations, we model the provider dynamics as the counterpart of user dynamics in Markov decision processes and RL. 

Second, while a single recommendation to a user only affects that specific user's utility, it can impact \emph{all relevant content provider} utilities. For instance, the providers who are recommended may become more satisfied with the platform, while those who are not can have their satisfaction decreased, especially if their content has not been shown to users for a while.
A recommendation objective that considers the utility of a single user and \emph{all} content providers may be intractable on platforms with large numbers of providers. To address this, we propose an approximation under which the all-content-providers objective can be relaxed to just the recommended content provider's \emph{utility lift}, defined as the difference in utility between the  provider being recommended versus not. 

We offer the following contributions:
\begin{itemize}
\item \textbf{RL for Content Provider Aware Recommendation}: We formulate recommendation as an RL problem for maximizing the joint objective of user and content provider utilities (Section \ref{sec:problem_formulation}). Building on the success of REINFORCE for user utility maximization \cite{chen2019top}, we propose a REINFORCE approach, coined \texttt{EcoAgent} (Section \ref{sec:recommender_design}), on top of a \textbf{neural sequential representation} of both user and content provider trajectories to maximize our joint objective.  
    \item \textbf{Counterfactual Content Provider Utility Uplift}: To address the computational concern of the effect of one recommendation on multiple content providers, we design a novel scalable objective that combines target user utility and recommended content provider utility \emph{uplift}; and demonstrate that it is equivalent to optimizing overall user and content provider utilities under a mild condition (Section \ref{sec:content provider-aware-objective}). 
    \item  \textbf{Multi-Stakeholder Simulation Environment}: To evaluate our approach in a setting where content providers' utilities are  affected by both the users' actions and the recommender actions, we build a Gym \cite{gym} environment (Section \ref{sec:environment}). This environment aims at capturing a simplified version of the user \emph{and content provider} dynamics, which is not supported in most existing simulation platforms.\footnote{Code of environment and RL algorithms can be found at \url{https://github.com/google-research/google-research/tree/master/recs_ecosystem_creator_rl}.} 
    \item \textbf{Study of User vs. Content Provider Trade-offs}: We provide a series of experiments, each with the goal of identifying the extent to which our content provider aware RL recommender can positively affect user satisfaction, and the overall multi-stakeholder health, as measured by the user vs.\ content provider utility Pareto curve, and the number of viable providers on the platform (Section \ref{sec:simulation}).
    
    
\end{itemize}

\section{Related Work}\label{sec:related_work}
The impact of policies on content providers has been actively studied in two-sided markets with consumers and providers. There are two important lines of research: one takes the game theory perspective and develops econometric models for marketplaces such as advertising auctions \citep{vickrey1961counterspeculation,edelman2007internet,athey2010structural}; another considers the fairness with respect to content providers in recommender systems \citep{biega2018equity,singh2018fairness,beutel2019fairness,mladenov2020optimizing,mehrotra2018towards}.  Table \ref{tab:related_work} summarizes the comparison between these works and ours.

These two lines of work often have different modeling focuses on  users and content providers, while our work aims at directly modeling their interaction dynamics with the platform. In the advertising auction literature, advertisers are modeled strategically \citep{vickrey1961counterspeculation,edelman2007internet,athey2010structural}. User (consumer) utility is implicitly incorporated in advertiser valuations. 
A similar gaming perspective is also introduced to platforms with strategic content providers  \cite{ghosh2014game,goel2012game,raifer2017information,ben2018game}. Conversely, literature focusing on fairness of recommender systems with respect to content providers often assumes fully observed or static provider states, based on which user utility and provider fairness are jointly optimized \citep{biega2018equity,singh2018fairness,mehrotra2018towards}.  However, as intelligent agents, content providers may have latent preferences/states influenced by users and recommender. To capture this, in contrast to prior work, we use Recurrent Neural Networks (RNNs) to model both user and content provider state transitions, and learn their sophisticated utility functions, which also implicitly account for their policies. With user and content provider characterizations in hand, we then update the agent through RL to optimize overall utilities.  

The idea of using RL while utilizing temporal information in recommender systems has received increasing attention. There is an emerging literature in neural recommender systems that applies RNNs to encode user historical events to predict future trajectories \citep{beutel2018latent,hidasi2015session,wu2017recurrent}. \citeauthor{chen2019top} utilize such a sequential represention as the user state encoding, on top of which an RL agent using a REINFORCE \citep{williams1992simple} policy-gradient-based approach is learned. Furthermore, the idea of jointly optimizing different objectives during agent learning can be categorized into a broader literature on {Multi-Objective Optimization} \cite{agarwal2011click, van2013scalarized, mossalam2016multi}.

Another difference of our work with the existing ones is the formulation of agent objective. In advertising auction literature, the goal is to construct an econometric model for a given marketplace, where advertiser strategies are inferred from the equilibria of the proposed model \citep{edelman2007internet,athey2010structural}. The platform generates rankings that trade off user satisfaction, platform revenue, and advertiser profits \citep{athey2010structural}.  We note a connection between advertiser strategies and our objective. A risk-neutral advertiser decides her bid by considering the incremental cost in an alternative ranking position \citep{edelman2007internet,athey2010structural}, while each content provider is considered by our \texttt{EcoAgent}  based on their incremental (uplifted) utility resulted from one recommendation.  

On the other hand, literature in fairness of recommender systems across content providers usually optimizes user satisfaction (ranking quality) as well as fairness with respect to content providers. The notion of fairness can be measured in multiple ways, such as content provider-groupwise model performance \cite{beutel2019fairness}, equity of content provider attention \cite{biega2018equity}, content provider popularity distribution  \cite{mehrotra2018towards}, content provider exposure \cite{singh2018fairness}, etc. As noted by \citeauthor{mehrotra2018towards}, such two objectives can be optimized through a combined metric of both (e.g., \cite{beutel2019fairness}) or constrained optimization, such as maximizing   fairness with having an acceptable ranking quality \cite{biega2018equity}, optimizing user utility while being fair to providers \cite{singh2018fairness}, etc. However, as pointed out by \citeauthor{asudeh2019designing}, metrics of fairness need careful design and are sometimes chosen subjectively. In contrast, our work directly maximizes overall content provider utilities and  implicitly considers disparity in content provider exposure and associated fairness in recommendations.

One closely related work featuring long-term `social welfare' optimization in recommender system is proposed by \citeauthor{mladenov2020optimizing}, wherein the authors optimize user utility while restricting  matched content providers to be viable in equilibrium \cite{mladenov2020optimizing}. Our work differs in a few major ways. First and foremost, we learn fully dynamic recommendation policies which adapt to user feedback and provider behavior (e.g. changing topic focus) in every time step in contrast to a precomputed query-specific policy. Moreover, due to the partially-observed nature of our setting, the \texttt{EcoAgent} recommender naturally factors in issues such as exploration and uncertainty representation.


Finally, we relate our work to literature applying \emph{counterfactual reasoning} to algorithm design, where the effect of a treatment (action, policy) is quantified by the difference of potential outcomes between the treated (with the treatment) and control (without the treatment) \cite{rubin1974estimating}. Usually, only one outcome is observed, and we need to infer the counterfactual alternatives. For example, \citeauthor{jaques2019social} define an influence reward to be the treatment effect of an agent's action on other agents, and use it to increase agent-level collaboration  in Multi-Agent RL \cite{jaques2019social}. \citeauthor{mehrotra2020inferring} apply it to music recommendation platforms, where they analyze the treatment effect of one artist new track release on popularities of herself  and other artists \cite{mehrotra2020inferring}. Similarly, in this paper, the content provider utility uplift is the treatment effect of current recommendation on content provider utility. 





\section{Problem Formulation}\label{sec:problem_formulation}
\begin{figure}[!t]
    \centering
    \includegraphics[width=0.85\linewidth]{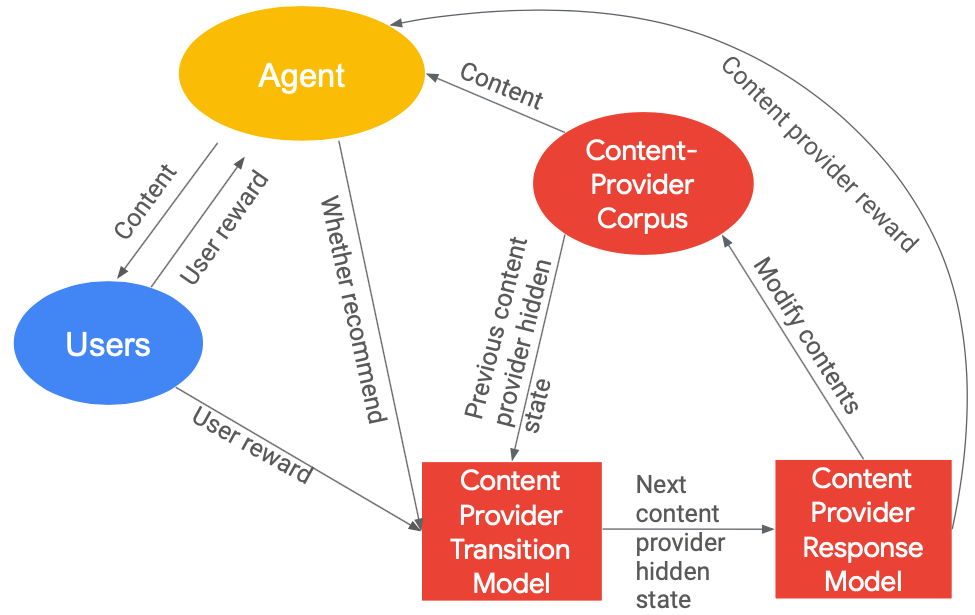}
    \caption{Interactions among Recommender Stakeholders}
    \label{fig:content provider_dynamics}
\end{figure}
We start by describing our problem setting, motivating the need for content provider aware RL-based recommenders and defining key terms to be used throughout.

\subsection{RL for User Utility}
\label{subsec:rl-user-utility}
\mmdel{While significant progress has been made in applying RL to recommendations \cite{chen2019top,strehl2010learning,joachims2017unbiased,swaminathan2017off}, most existing work focuses on  learning recommendation policies to maximize long-term \emph{user utility}.}

A typical setup for long-term recommendation problems is as follows \cite{chen2019top,strehl2010learning,joachims2017unbiased,swaminathan2017off}. A user issues a (possibly implicit) query to the recommendation system. The recommender agent responds with a slate of one or more items of content; the user interacts with the recommended content according to their (unobserved) preferences, emitting signals, such as likes, clicks, dwell times, indicating their engagement, satisfaction, etc. These proxy reward signals are then collected by the agent for future  planning. We use \emph{proxy reward} to refer to these immediate signals indicative of the user's affinity towards the recommended content, which can depend on topic preference alignment and content quality, among others. 

We formulate the problem as a Markov decision process (MDP)\footnote{Recommender systems are generally best formulated  as \emph{partially-observable MDPs} since user states are not observed. Here, for ease of exposition, we retain the MDP notation, assuming a learned state representation; but throughout (including experiments), we \emph{infer} a user's state from their interaction history with the platform.} $(\mathcal{S}^u, \mathcal{A}, P^u, R^u, \rho_0^u, \gamma^u)$, where $\mathcal{S}^u$ is the state space encoding user topic preferences and context, $\mathcal{A}$ is the action space composed of the content available on the platform\footnote{W.l.o.g., we assume the agent recommends a single item to the user each time.}, 
$P^u: \mathcal{S}^u\times \mathcal{A} \rightarrow \Delta(\mathcal{S}^u)$
captures the state transition, $R^u$ defines the user proxy reward, $\rho_0^u$ is the initial state distribution, and $\gamma^u$ the discount factor. A recommendation policy $\pi(\cdot|s)$ over the action space is learned to maximize the discounted cumulative reward across different user trajectories, which we refer to as the expected \emph{user utility} 
 under the policy $\pi$, that is
\begin{eqnarray}
\label{eq:user_rl}
    \max_\pi && \sum_{t=0}^{T} \mathbb{E}_{s^u_t \sim d_{\pi, t}^u(s_t^u), a_t\sim \pi(\cdot|s^u_t)} \left[Q_t^u(s^u_t,a_t)\right]\\
    \mbox{where} && Q^u_t(s^u_t,a_t)=\sum_{t'=t}^{|\tau^u|} \left(\gamma^u\right)^{t'-t}r^u(s^u_{t'}, a_{t'})\nonumber .
\end{eqnarray}
Here $T$ is the maximum trajectory length, and $\tau^u$ is the user trajectory sampled according to the policy $\pi$ under the user MDP. 
We use $d_{\pi, t}^u(\cdot)$ to denote the average user state visitation distribution at time $t$ \cite{levine2020offline}, and $Q^u(s^u_t,a_t)$ is the user utility calculated from time step $t$ according to the policy $\pi$. Note that the expectation takes into account different user trajectories induced by the learned policy.

\subsection{Multi-Stakeholder Interactions}
\label{subsec:interactions}
The above framing misses one group of key stakeholders: the content providers. To be able to optimize the overall utility of \emph{all} stakeholders, it is necessary to extend the user-recommender interaction model to account for the impact of recommendations on providers' state and utility. Figure~\ref{fig:content provider_dynamics} gives an intuitive schema that captures stylized, yet real-world-inspired interactions among these stakeholders. As a simplification, we do not include environment externalities, e.g., exogenous discovery of provider. 

\subsubsection*{Users $\leftrightarrow$ Recommender Agent} The interactions between users and the recommendation agent are already captured in the setup of the previous section. One point to highlight is that not only does the agent adapt to user feedback, but users themselves are influenced by their interactions with the platform (e.g., a user's topic preferences might shift closer to or further from recommended content topics depending on how much they like the recommendations).

\subsubsection*{Content Providers $\leftrightarrow$ Recommender Agent} Besides users, the agent also influences the content providers.
This form of interaction has been largely ignored in the recommendation literature, but it is a key focus of this work.  Specifically,  depending on the exposure generated by the platform (i.e., how much the agent recommends a provider's content to users) and the user feedback (the users' \emph{proxy reward} for such recommended content), content providers may decide to change the nature of their engagement with the platform. 
For example, content providers can decide to create more or less content, shift their topic focus, and or even leave the platform if their satisfaction falls below a certain point. 
This in turn influences the recommender agent, as it will directly affect the content corpus, from which recommendable candidate items are drawn. 


\subsection{RL for Content Provider Utility}
\label{sec:rl_content provider}


Similar to a recommender agent that maximizes user utility as defined in Section~\ref{subsec:rl-user-utility}, we can also define a MDP $(\mathcal{S}^c, \mathcal{A}, P^c, {R}^c, \rho_0^c, \gamma^c)$ for each content provider $c$. Here $\mathcal{S}^c$ is the content provider state space, $\mathcal{A}$ is again the action space composed of the content available on the platform, and
${P}^c: \mathcal{S}^c\times \mathcal{A} \rightarrow \Delta(\mathcal{S}^c)$
captures the content provider state transition, ${R}^c$ defines the content provider proxy reward, $\rho_0^c$ is the initial state distribution and $\gamma^c$ the discount factor. 
We explain the content provider state, transition model and proxy reward in more detail.

 \textbf{Content Provider State $s^c_t$}: 
 The content provider state is influenced by the popularity of their current content, and captures the  
 latent incentives for future content creation and preferences for content topics. 
   In our design, we also force one component of the state to capture the content provider's current satisfaction with the platform, specifically the cumulative
   historical 
   content provider proxy reward up to time $t$.

    \textbf{Content Provider State Transition $s^c_{t+1} \sim P^c(\cdot|s^c_t, a_t)$}: Content provider states are changing over time based on feedback from the recommender and the users, i.e., how many recommendations the content provider received and how the users liked this content provider's recommended content. For instance, if content provider's existing content is popular among users, the content provider's latent topic preference will shift towards existing content topics; otherwise they may change topics of future creation. The content provider satisfaction component of the state is incremented by the number of recommendations and summed user reward proxy signals acquired from the current time step. 
    
     \textbf{Content Provider Reward Signals $r^c(s_t^c, a_t)$ and Utility}: We use the content provider \emph{proxy reward} $r^c$ to capture content providers' incremental engagement/satisfaction with the platform at any time point. Such signals can manifest as viability (i.e., decide to stay/leave the platform), frequency of new content uploads, etc.\footnote{If we view content providers as rational decision-making agents, which could in turn be modeled as RL agents themselves,  whether they will leave the platform, what type of content they create, etc., can be seen as content provider actions rather than reward proxy signals. Here though, we model this from the perspective of recommender as the RL agent finding the action so to maximize content provider--and user--utility.} Conceptually, content provider reward captures how active the content provider is during a given time period. 
    It also indicates how the content provider responds to  \emph{recommender and user feedback}. We argue that content providers of different groups will respond differently.  Particularly, \emph{less established content providers} who receive little attention or negative user feedback during certain time period are more likely to change their next content topics or become non-viable and eventually leave the platform, while the \emph{established content providers} are more resilient to small changes in recommender and user feedback. 

With the content provider MDP, we can learn a recommendation policy $\pi(\cdot|s)$ over the action space to maximize the discounted cumulative reward, which we refer to as the expected \emph{content provider utility} 
under the policy $\pi$, that is 
\begin{eqnarray}
\label{eq:creator_rl}
    \max_\pi&& \sum_{t=0}^{T}\mathbb{E}_{s^c_t\sim d_{\pi, t}^c(s^c_t), a_t\sim \pi(\cdot|s^c_t)} \left[Q^c_t(s^c_t,a_t)\right]\\
    \mbox{where} && Q^c_t(s^c_t,a_t)=\sum_{t'=t}^{|\tau^c|} \left(\gamma^c\right)^{t'-t}r^c(s^c_{t'}, a_{t'})\nonumber .
\end{eqnarray}
Here $T$ is the maximum trajectory length, and $\tau^c$ is the creator trajectory sampled according to the policy $\pi$ under the creator MDP. Similarly, we use $d_{\pi, t}^c(\cdot)$ to denote the average content provider state visitation distribution at time $t$ \cite{levine2020offline}, and $Q^c(s^c_t,a_t)$ is the content provider utility calculated from time step $t$ according to policy $\pi$. Again the expectation takes into account the different content provider trajectories induced by the learned policy.

\subsection{RL for Content Provider and User Utility}
\label{sec:content provider-aware-objective}
Having established that recommendations 
affect the future state and utility of both users \emph{and content providers}, we formulate the recommendation problem as follows: 
\begin{quote}
\emph{Decide which content (from which content provider) to recommend to a user so that a combined metric of both content provider and user utilities is maximized, given the current state of the user and the providers of candidate content recommendations.} 
\end{quote}

Putting everything together, we translate this setup into a user-and-content-provider MDP $(\mathcal{S}, \mathcal{A}, {P}, {R}, \rho_0, \gamma)$ where $\mathcal{S}$ is the concatenation of user state and states of all the content providers on the platform, $\mathcal{A}$ is the content available on the platform,
${P}: \mathcal{S}\times \mathcal{A}  \rightarrow \Delta(\mathcal{S})$ 
captures the state transition, $R$ is the concatenation of proxy rewards of the user and all content providers, $\rho_0$ is the initial state distribution, and $\gamma$ as concatenation of user and content provider discount factors. With this user-and-content-provider MDP, our goal is to learn a recommendation policy $\pi(\cdot|s)$ over the action space that maximizes two objectives: user utility and content provider utility. For simplicity, we adopt a scalarization approach, using  a coefficient $\lambda \in [0, 1]$ to interpolate between the two objectives, which we refer to as the \emph{content provider constant}; future work can look into other multi-objective optimization (MOO) approaches as well (e.g., \cite{agarwal2011click, van2013scalarized, mossalam2016multi}).
When $\lambda=0$, this is equivalent to RL for user utility (Section~\ref{subsec:rl-user-utility}), whereas when $\lambda=1$, this is a content provider-only optimizing policy (Section~\ref{sec:rl_content provider}).\footnote{Nevertheless, as content provider reward proxy signals are a function of user feedback, this still indirectly optimizes user utility; but this might not always be the case depending on how one decides to define content provider proxy reward.} 

\begin{eqnarray}
\label{eq:joint-obj}
\max_\pi \quad \sum_{t=0}^{T}\mathbb{E}_{\substack{s_t\sim d_{\pi, t}(s_t) \\ a_t\sim \pi(\cdot|s_t)}}\left[ (1-\lambda) Q^u_t(s^u_t,a_t) + \lambda \sum_{c\in\mathcal{C}} Q^c_t(s^c_t,a_t)\right] 
\end{eqnarray}
 where $T$ is the maximum trajectory length, $\mathcal{C}$ is the set of content providers on the platform\footnote{W.l.o.g., we assume that all content providers have the same discount factor.}, state $s_t=(s^u_t, s^{c_1}_t,\dots, s^{c_{|\mathcal{C}|}}_t)$ is the state concatenation of the user and all content providers, $d_{\pi, t}(s_t)$ is the average visitation distribution of $s$ at time $t$ \cite{levine2020offline}, $Q^u_t$ and $Q^c_t$ are user and content provider utilities at time $t$ defined in Eq.~\ref{eq:user_rl} and Eq.~\ref{eq:creator_rl} respectively. Note that for each recommendation (user query), we consider the utility of the targeted user and the utilities of all relevant content providers in the action space---the content provider who is recommended increases her satisfaction, while others who are not have their satisfaction decreased.

Albeit complete, the objective as defined in eq.~\ref{eq:joint-obj} is intractable on platforms with a large number of  content providers. To make it tractable, we break down the utility of the content providers into the one whose content was recommended by the recommender and the others. That is, 
\begin{eqnarray}
\label{eq:joint-obj-breakdown}
       \sum_{c\in\mathcal{C}} Q^c_t(s^c_t,a_t) &=&  Q_t^{c_{a_t}}(s^{c_{a_t}}_t, a_t) + \sum_{c'\in\mathcal{C}\setminus c_{a_t}} Q_t^{c'}(s^{c'}_t, a_t)
\end{eqnarray}
Here we use $c_{a_t}$ to denote the content provider associated with the chosen content $a_t$.

\begin{assumption}[No Content Provider Externality]\label{assump:no_content provider_externality}
A content provider's proxy reward and state transition of not being recommended do not depend on which provider is recommended instead. 
\end{assumption}
Intuitively, the above assumption describes that when a provider is not recommended, her reaction  only depends on the fact that she  was not exposed to users, but not on which other providers were recommended. We translate it to  the following two equivalences:  for any content provider $c'\in \iC\setminus c^{a_t}$ who is not chosen for current recommendation, we have
\begin{eqnarray}
   \mbox{\emph{proxy reward} } && r^{c}(s^{c'}_t, a_t) = r^{c}(s^{c'}_t, b^{c'}_t) \nonumber \\
   \mbox{\emph{state transition}} && P^{c}(\cdot| s^{c'}_t, a_t)\stackrel{d}{=} P^{c}(\cdot| s^{c'}_t, b^{c'}_t)\nonumber 
\end{eqnarray}
where $b^{c'}_t$ indicates any chosen content not associated with the content provider $c'$ at time $t$. This further implies that, for any content provider $c'\in \iC\setminus c^{a_t}$ who is not recommended  at current time step, her utility later on  does not depend on which  other provider's content is recommended, i.e.,
\begin{equation}
    Q^{c}_t(s^{c'}_t, a_t)=Q^{c}_t(s^{c'}_t, b^{c'}_t),\quad \forall c'\in  \iC\setminus c^{a_t}.
\end{equation}
Eq.~\ref{eq:joint-obj-breakdown} can then be simplified as the recommended content provider \emph{utility uplift} plus a baseline as the sum of all content provider utilities when none of their content are recommended,
\begin{eqnarray}
\label{eq:joint-obj-breakdown2}
  &&\sum_{c\in\mathcal{C}} Q_t^c(s^c_t,a_t) \\
  = && \underbrace{Q_t^{c_{a_t}}(s_t^{c_{a_t}}, a_t) -Q_t^{c_{a_t}}(s_t^{c_{a_t}}, b^{c_{a_t}}_t)}_{\mbox{\emph{utility uplift}}} +\underbrace{ \sum_{c'\in\mathcal{C}} Q_t^{c'}(s^{c'}_t, b^{c'}_t)}_{\mbox{\emph{baseline}}}.\nonumber 
\end{eqnarray}
Note that the baseline has the merit of independence from the taken action and thus can be dropped when calculating the gradient of policy~\citep{williams1992simple}, which will be detailed in Section \ref{sec:recommender_design}. As a result, the objective defined in eq.~\ref{eq:joint-obj} can be simplified as optimizing user utility and the utility uplift of the recommended content provider,
\begin{eqnarray}
\label{eq:joint-obj-simplified}
    \max_\pi& & \sum_{t=0}^T\mathbb{E}_{s_t\sim d_{\pi, t}(s_t), a_t\sim \pi(\cdot|s_t)}[R_t]\\
    \mbox{where} && R_t= (1-\lambda) Q^u_t(s^u_t,a_t)\nonumber  \\
    &&\quad\quad+ \lambda \left( Q_t^{c_{a_t}}(s^{c_{a_t}}_t, a_t) - Q_t^{c_{a_t}}(s_t^{c_{a_t}}, b^{c_{a_t}}_t)
    \right).\nonumber
\end{eqnarray}

\begin{figure*}
    \centering
    \includegraphics[width=0.8\linewidth]{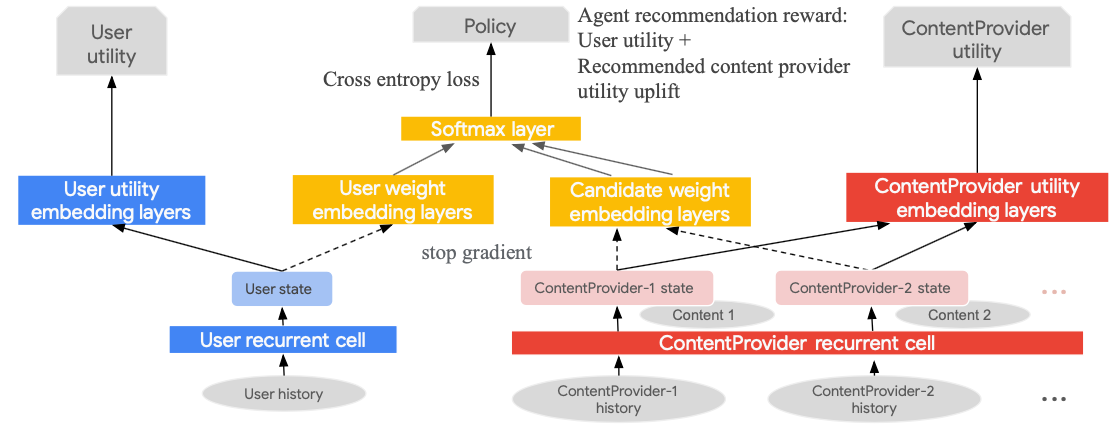}
    \caption{Illustration of \texttt{EcoAgent} structure. \texttt{EcoAgent} consists of three components: (i) a user RNN utility model that embeds user history into user hidden states and predicts user utility; (ii) a content provider RNN utility model that embeds content provider history into content provider hidden states and predicts content provider utility; (iii) an actor model that inputs user hidden state and candidates (content, content provider hidden state) to generate policy. Actor model is optimized using REINFORCE with recommendation reward being a linear combination of user utility and content provider utility uplift. 
    }
    \label{fig:EcoAgent}
\end{figure*}
 
\section{Provider-Aware REINFORCE  Agent}\label{sec:recommender_design}
Now that we have formulated the multi-stakeholder recommender problem as an RL problem, we turn attention to learning the policy $\pi$ that maximizes the objective Eq.~\ref{eq:joint-obj-simplified}. Different approaches~\citep{mnih2013playing,van2015deep,sutton2000policy,levine2013guided,williams1992simple} can be applied to solve this RL problem.
Here, we adopt the policy-based REINFORCE \citep{williams1992simple} approach, inspired by its success in the classic user utility maximization recommender setting \cite{chen2019top}. 

Let (parameterized) policy $\pi_\theta$ map states to actions. Using the log-trick, the policy parameters $\theta \in \mathbf{R}^d$ can be updated directly following the gradient,  
\begin{eqnarray}\label{eq:reinforce}
  \mathbb{E}_{s_t\sim d_{\pi, t}(s_t), a_t\sim \pi(\cdot|s_t)} \left[R_t\log\nabla_{\theta}\pi_{\theta}(a_t|s_t) \right].
\end{eqnarray}
Eq.~\ref{eq:reinforce} gives an unbiased estimate of the policy gradient in online RL, where the gradient of the policy is computed on the trajectories collected by the same policy; this is our setup here. 

In what follows, we outline how we parameterize the policy---particularly how to learn the user states (Section \ref{subsec:user-utility}), the content provider states (Section \ref{subsec:content provider-utility}), and the policy-specific parameters on top of these states to output probabilities over the action space (Section \ref{subsec:policy-softmax}). Finally, we discuss in detail our reward, and specifically how we learn the content provider utility uplift part of the reward (Section \ref{subsec:reward}), and close this section with an overview of how everything ties together to solve the  multi-stakeholder recommendation RL problem (Section~\ref{subsec:nutshell}). We refer to our proposed learning approach as \texttt{EcoAgent}, illustrated in Figure~\ref{fig:EcoAgent}.

 \subsection{User States via Utility Imputation}
  \label{subsec:user-utility}
As described in Section~\ref{subsec:rl-user-utility}, the user state $s^u \in \mathcal{S}^u$ encodes user preferences and context.  To learn the user state, building on recent work showing the value of sequence learning \cite{chen2019top}, we use a recurrent neural network (RNN) to encode user interaction histories, and ensure the learned user state together with the action representation can predict observed user utility. 

Let a user trajectory 
$\tau = \{(x_0^u, a_0, r^u_0), \cdots, (x_T^u, a_T, r^u_T)\}$ sampled according to the current policy $\pi_\theta$, where $x_t^u$ is user's  context at time t. 
We can then further break $\tau$ into a set of history-action-return tuples of $\mathcal{D}^u = \{(H^u_t, a_t, Q^u_t)\}$ by aggregating the historical contexts and actions $H^u_t=\{(s_{t'}^u, a_{t'})\}_{t'=0}^{t}$ and accumulating the discounted future reward $Q^u_t$.

Given this collection $\mathcal{D}^u$ of historical contexts/actions and associated utilities, we can learn user utility model parameters $\phi$ so to  minimize a loss function $\ell$ between the predicted and ground truth user utilities. This also  offers the user hidden state $s^u_t$ at time $t$:
\begin{eqnarray}
    \min_{\phi}\sum_{(H^u_t, a_t, Q^u_t) \in \mathcal{D}^u} \ell\left(\hat{Q}_t^u(s_t^u, a_t; \phi), Q^u_t\right)\\
   \mbox{where } s_t^u = \mbox{RNN}(H^u_t)\nonumber.
\end{eqnarray}
In our experiments, we used Huber loss for $\ell$. 

 
  \subsection{Provider States via Utility Imputation}
  \label{subsec:content provider-utility}
  As described in Section \ref{sec:rl_content provider}, the content provider states encode content provider preferences and satisfaction. We learn the content provider states in similar fashion as the user states, ``forcing'' the state to be able to predict content provider utility. 
  
Given a provider trajectory $\tau = \{(x_0^c, A_0, r^c_0), \cdots, (x_T^c, A_T, r^c_T)\}$, where $x_t^c$ is content provider $c$'s context at time $t$,
we again break it into a set of history-action-return tuples of $\mathcal{D}^c = \{(H^c_t, A_t, Q^c_t)\}$ by aggregating historical contexts and actions, and accumulating the discounted future reward. In general, we expect content provider dynamics to evolve at a much slower rate than user dynamics; hence, the per-event time interval will typically be much different from that of the user trajectory. 
At each time step, the action $A_t = \{(v^{d_i}, r^u_i)\}_{i=1}^{m_t}$ records the provider's recommended item $d_i$ and the user feedback $r^u_i$ offered on this content, where $m_t$ denotes the number of recommendations made of this provider's contents, and $v^{d_i}$ as an indicator vector of content $d_i$. 
Concretely, $A_t$ is summarized by (i) the number of recommendations $m_t$ this provider received at time $t$; (ii) the sum of user rewards $\sum_{i=1}^m r_i^u$; and (iii) a weighted bag-of-words representation of the recommended contents $\{d_i\}_{i=1}^{m_t}$, with weights reflecting the received user rewards: $\sum_{i=1}^{m_t} r_iv^{d_i} /\sum_{i=1}^{m_t} r_i $.
 
We employ another RNN to encode the historical actions into a content provider's state and learn the model parameters $\psi$ so to minimize a loss function $\ell$ between the predicted  and the ground truth content provider utilities, 
\begin{eqnarray}
\label{eq:content provider-state-utility}
    \min_{\psi}\sum_{(H^c_t, A_t, Q^c_t) \in \mathcal{D}^c} \ell\left(\hat{Q}_t^c(s_t^c, A_t; \psi), Q^c_t\right)\\
   \mbox{where } s_t^c = \mbox{RNN}(H_{t}^c)\nonumber.
\end{eqnarray}

  \subsection{Policy Parameterization}
  \label{subsec:policy-softmax}
  Conditioning on the user state $s^u$ as learned by the User Utility model (Section~\ref{subsec:user-utility}) and each candidate content provider states $s^c$  as learned by the content provider utility model (Section~\ref{subsec:content provider-utility}), the policy $\pi_\theta (a|s)$  mapping state to action is then learned with a simple softmax,
   \begin{equation}
    \pi_\theta (a|s) = \frac{\exp\langle \Phi({s^u}), \Psi([a, s^{c_a}])\rangle/T)}{\sum_{a' \in \mathcal{A}} \exp(\langle \Phi({s^u}), \Psi([a', s^{c_{a'}}])\rangle/T)},
\end{equation}
Note that this follows the paradigm of latent factor based recommendation where both users and candidate content items are mapped to the same low-dimensional embedding space, and the inner product $\langle, \rangle$ is used to capture the user-content proximity. Here, we use $\Phi$ to denote the projection of user state $s^u$ to the embedding space. For the representation of the candidate content item, we concatenate the dense action/content embedding $a$ with its associated content provider state $s^{c_a}$, and map them, via $\Psi$ into the same low-dimensional space. 
$T$ is a temperature term controlling the smoothness of the policy (set to $1$ in our case).



\subsection{Reward}
 \label{subsec:reward}
Last, we need to specify our reward which will be guiding the learning of the policy. Recall that in eq.~\ref{eq:joint-obj-simplified} we have defined our reward as the $\lambda$-weighted sum between the user utility and the content provider utility uplift.

 \paragraph{User Utility Reward} Specifically, for user $u$ and  recommendation $a_t$ at time $t$, we calculate user utility $Q_t^u(s^u_t, a_t)$  using  the accumulated $\gamma^u$-discounted future reward of this user:
 \begin{equation}
     {Q}_t^u(s^u_t, a_t) = \sum_{t'=t}^\infty  \left(\gamma^u\right)^{t'-t} r^u(s_{t'}^u, a_{t'}).
 \end{equation}
 
 
 \paragraph{Content Provider Utility Uplift Reward} For the recommended content provider $c^{a_t}$, recall that this is defined as the difference of content provider factual utility when recommended versus the counterfactual utility when not recommended, as derived in eq.~\ref{eq:joint-obj-breakdown2} and shown below,
 \begin{equation}
 \label{eq:content provider_uplift_utility}
     Q_t^c(s_t^{c_{a_t}},a_t) - Q_t^c(s_t^{c_{a_t}}, b_t^{c_{a_t}})
 \end{equation}

\begin{figure}
    \centering
    \includegraphics[width=\linewidth]{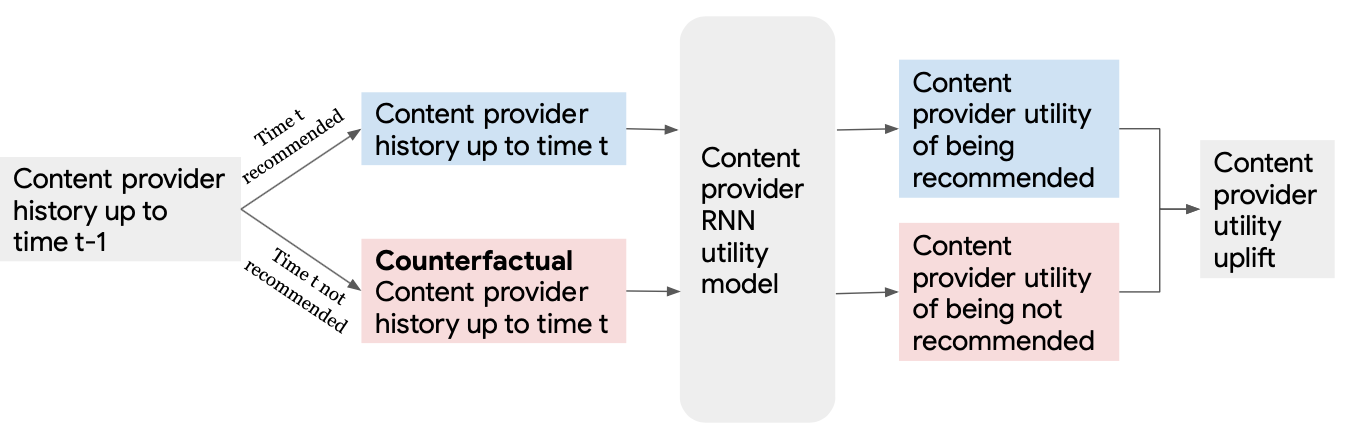}
    \caption{Illustration of Content Provider Utility Uplift. 
    }
    \label{fig:content provider_uplift_utility}
\end{figure}

 However, for provider $c_{a_t}$, we do not have the data of her counterfactual utility supposing that she was not recommended at time $t$. We thus resort to content provider RNN utility model $\hat{Q}^c(s_t^c, A_t'; \psi)$  in Eq.~\ref{eq:content provider-state-utility} to predict her counterfactual utility value. Here $A_t'$ stands for the counterfactual history where we remove the chosen action $a_t$ from this content provider. To be robust to systematic prediction bias, we also use the 
 utility model to estimate her factual utility for consistency. We illustrate this calculation in Figure~\ref{fig:content provider_uplift_utility}.  

  \subsection{Putting Everything Together}
  \label{subsec:nutshell}
  We adopt online learning for training \texttt{EcoAgent} and leave the discussion on offline learning to Section  \ref{sec:discussion}. In online RL, agent interacts with the environment to collect new data, based on which the agent is updated. We thus use the following optimization procedure: \begin{enumerate}[(i)]
      \item (\texttt{EcoAgent} in Action) Use current parametrized policy $\pi_\theta$ to interact with the environment and collect new data;
      \item (\texttt{EcoAgent} in Optimization) Use newly-collected data to optimize the policy $\pi_{\theta}$, and particularly the parameters of the user utility RNN model $\phi$, the parameters of the content provider utility RNN model $\psi$, and the policy softmax parameters;
      \item Repeat (i) and (ii) until converged.
  \end{enumerate}

\section{Simulated Environment}
\label{sec:environment}

We opt for a simulation environment to study the effectiveness of our approach for two reasons. First,
publicly available recommender datasets do not consider the content provider aspect of the recommendation problem; they therefore cannot serve as a basis for evaluating RL for multi-stakeholder recommendation. Second, simulation allows us to control the underlying environmental assumptions and study their effects on the performance of our proposed approach. 

Although there has been prior work on simulation recommendation environments \cite{mladenov2020optimizing}, it does not capture the full spectrum of users-recommender-content providers interactions described in Section~\ref{sec:problem_formulation}. 
We thus implemented a new Gym \cite{gym}  environment on top of RecSim which is a simulation platform with user dynamics \citep{ie2019recsim}. This environment is designed to capture the content provider dynamics as illustrated in Figure \ref{fig:content provider_dynamics}. 
We now offer more detail on the underlying dynamics used in our experiments. 

\subsection{Content Description}
\label{subsec:content}
We assume that there are $K$ content \emph{topics}. Each content item $d$ has an observable one-hot vector $v^d\in \{0,1\}^K\sim \mbox{Categorical}(p)$ representing its topic and a scalar \emph{quality} $q^d\sim \mbox{Truncated normal}$$(\mu$$, \sigma$,$ -1$$, $$1)$ that is perceived by the user but is hidden to the agent. Both $p$ and $(\mu,\sigma)$ are content provider-dependent, with $p$ reflecting a provider's topic preference of future creation, which can shift during the provider's interaction with the platform. 

\subsection{User Updates}
Each user $u$ is described by a unit vector $v^u_t \in \iR^K$ that represents her topic preference at time $t$. The initial state $v^u_0$ is sampled from a uniform distribution over the unit ball. After being recommended some item, the user immediate reward $r^u_t$  is a linear combination of the content's relevance (defined as the inner product of user topic preference $v^u_t$ and content topic $v^d$) and  content quality $q^d$, with $\eta^u$ denoting the user's sensitivity to content quality: $r^u_t=(1-\eta^u)\langle v^d, v^u_t\rangle + \eta^uq^d$. User topic preference then shifts toward the topic of the recommended content, weighted by how much the user liked it, $r^u$; thus $v^u_{t+1} \leftarrow v^u_{t} + \delta^u r^u v^d$, with $\delta^u$ denoting the degree to which the user's topic preferences is influenced by the recommended content.

\subsection{Content Provider Updates}

Each content provider $c$ is described by a vector $v_t^c \in \iR^K$ that represents her topic preference for future creation at time $t$. The initial state $v_0^c$ is  sampled from a uniform distribution over the unit ball. 
Each provider starts with a fixed number of content items, but the amount of content can increase over time if she decides to create more. 

Suppose that at time $t$, provider $c$ receives $m$ recommendations of her content $(d_1, \dots, d_m)$  and user rewards $(r^{u_1}, \dots, r^{u_m})$. We aggregate the recommender-induced exposure and user feedback as
\begin{equation}
\label{eq:popularity_feedback}
p_t^c = \mu^c + \eta^c_1m + \eta^c_2\sum_{i=1}^m r^{u_m},
\end{equation}
where $\eta^c_1, \eta^c_2$ represent content provider sensitivity  to content exposure and user feedback respectively, and $ \mu^c<0$ accounts for the \emph{negative impact of no recommendation}. Provider satisfaction at time $t$ is then defined as
\begin{equation}
    S^c_t = f\left(\sum_{s=0}^t p_s^c\right).
\end{equation}
Here $f$ transforms the raw feedback into content provider satisfaction. Different choices of $f$ can represent different content provider characteristics, which we detail below.


Content provider reward is then defined to be incremental provider satisfaction:
\begin{equation}
    r^c_t = S^c_t - S^c_{t-1}.
\end{equation}
Specifically, when $f$ is the identity, the provider reward is simply the raw feedback $p_t^c$. If her current reward $r_t^c$ is positive, the provider will create more content, where the number of new items is proportional to $r_t^c$.

To model content providers adapting their topic preferences based on user feedback, a content provider updates her topic preference using the sum of her recommended content topics, weighted by the corresponding user feedback: 
\begin{equation}
v^c_{t+1} \leftarrow v^c_{t} + \delta^c\sum_{i=1}^m r^{u_i}v^{d_i},
\end{equation}
where $\delta^c$ represents how sensitive the content provider's topic preferences are to the user feedback. 

Each content provider also has a viability threshold; if their satisfaction $S^c_t$ is below the threshold, they leave the platform. 

\subsubsection*{Content Provider Satisfaction Design}
\begin{figure}
    \centering
    \includegraphics[width=\linewidth]{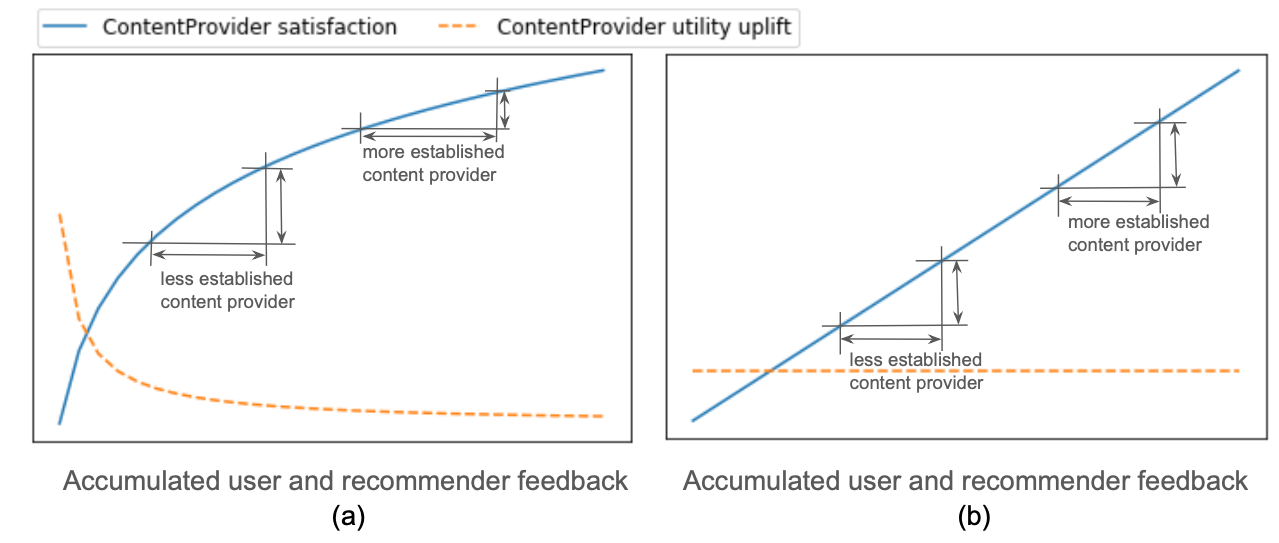}
    \caption{Two configurations for content provider satisfaction function $f$ over accumulated recommender and user feedback: (a) log function, saturating as recommender and user feedback gets increased, (b) linear function, where there is no difference of recommendation effect on content provider utilities between a less established content provider versus an established content provider.}
    \label{fig:linear-log-utility}
\end{figure}

As mentioned, content provider dynamics are characterized by the function $f$, transforming the raw recommender and user feedback to content provider satisfaction. While a linear transformation is the most intuitive, concave functions such as logarithm (see Figure~\ref{fig:linear-log-utility}a), reflect  the law of diminishing returns often seen in real life. For example, as discussed in Section~\ref{sec:rl_content provider}, one recommendation may not influence established content providers much given their existing popularity base. In contrast, for less established providers who are considering leaving the platform due to insufficient prior exposure, a new recommendation may change their mind and thus help the platform retain them. With a concave satisfaction function, less established content providers will 
derive more utility
than more established providers given the same amount of raw recommender and user feedback $p_t^c$. 
We often adopt this general assumption:
\begin{assumption}[Saturated content provider satisfaction]
\label{assump:saturared_content provider}
Content provider satisfaction gets saturated with accumulated recommender and user feedback. 
\end{assumption}




\section{Experiments} \label{sec:simulation}
After defining our simulation environment, we now turn to our experiment results.







\subsection{Experiment Setup}
\begin{figure*}
    \centering
    \includegraphics[width=0.7\linewidth]{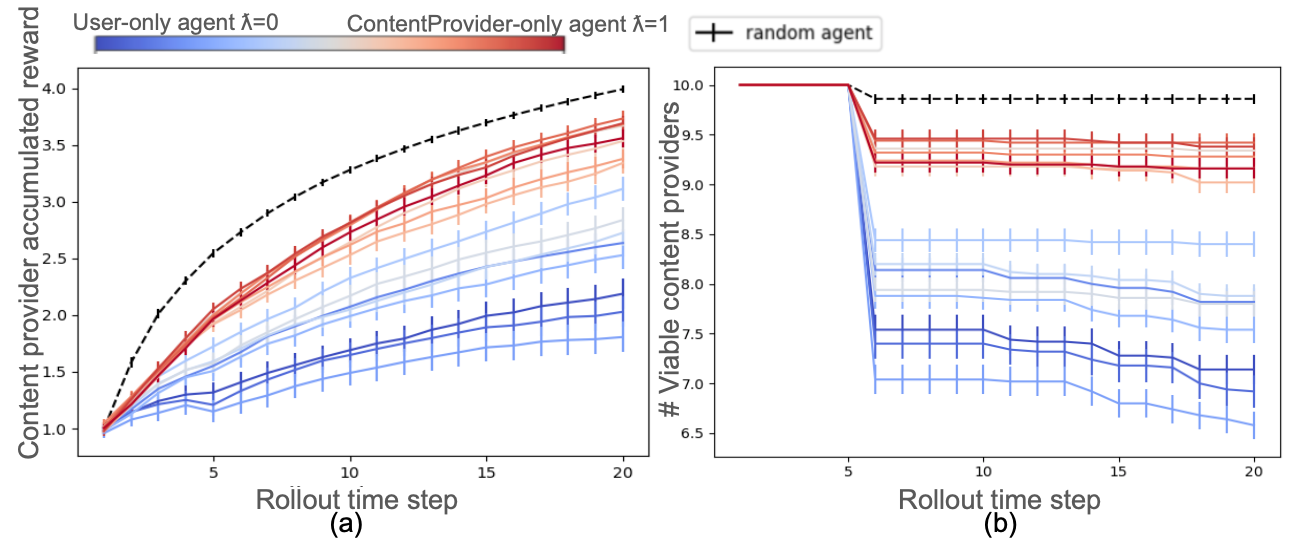}
    \caption{A content provider oriented \texttt{EcoAgent} ($\lambda$ close to $1$) helps content providers by improving content provider accumulated reward and number of viable content providers as compared to a user-oriented \texttt{EcoAgent} ($\lambda$ close to $0$).}
    \label{fig:submodular_content provider_help}
\end{figure*}


During online learning, \texttt{EcoAgent} collects data by interacting with a new environment with prespecified hyperparameters characterizing user and content provider dynamics described in Section \ref{sec:environment}. Each new environment samples initial states of $50$ users and $10$ content providers uniformly from the state spaces. Each content provider is initialized with $20$ content items which are sampled from $10$ topics based on provider's topic preference\footnote{We design our environment to have  more users than content providers, and the number of items is larger than the number of users, following the statistics on current content sharing platforms.}, and they have the choice of creating more content or leaving the platform as described in Section \ref{sec:environment}. New comers are not considered. 

For each training epoch, \texttt{EcoAgent} interacts with $10$ new environments as set up above. The environment  will be rolled out for $20$ steps. At each time step of one rollout, all users receive recommendations simultaneously, and the environment updates all users' and content providers' states.  We then use collected data to update \texttt{EcoAgent} with Adagrad optimizer. Throughout the experiments, we consider  long-term effects on users and content providers by setting discount factors: $\gamma^u=\gamma^c=0.99$. 

\subsubsection{Baselines} For each environment setup (user and content provider dynamics), we consider \texttt{EcoAgent}s with different $\lambda$'s (content provider constant)  varying from $0$ to $1$. We highlight three baselines:
\begin{itemize}
    \item \textbf{user-only \texttt{EcoAgent}}, where $\lambda=0$. \texttt{EcoAgent} only optimizes user utility;
    \item \textbf{content provider-only \texttt{EcoAgent}}, where $\lambda=1$. \texttt{EcoAgent} only optimizes content provider utility;
    \item \textbf{random agent}, where the agent recommends content randomly from the candidate set of content items.
\end{itemize}
Conceptually, user-only \texttt{EcoAgent} should have the largest user satisfaction, while content provider-only \texttt{EcoAgent} has the largest content provider satisfaction. The random agent sets a baseline on how \texttt{EcoAgent} learns user preference as well as on how \emph{content provider aware} ($\lambda>0$) \texttt{EcoAgent}s help content providers. 

For every agent considered, we use LSTM cells in utility models with $(32,32,16)$ hidden layers between LSTM output and utility prediction, which are chosen by preliminary experiments to minimize the loss of utility prediction. For actor model, we first embed user states and candidate (content provider state, content) tuples into weights using $(32,32,32)$ fully-connected layers (again tuned by preliminary experiments), then generate policy based on the softmax of the dot product between user weight and candidate weights.  We tune \texttt{EcoAgent}s with various learning rates and  select the model that maximizes the objective function eq.~\ref{eq:joint-obj-simplified} with corresponding content provider constant $\lambda$. 


\subsubsection{Metrics} To compare different agent models, we test them in $50$ rollouts of new environments for  $20$ steps (with prespecified user and content provider dynamics but different initial states). We  calculate the following statistics of each rollout, which  summarize how the agent influences users and content providers respectively:
\begin{itemize}
    \item \textbf{user accumulated reward}: $\sum_{t=1}^{20} r_t^u$;
    \item \textbf{content provider accumulated reward}: $\sum_{t=1}^{20} r_t^c$;
    \item \textbf{\# viable content providers}: number of providers in the environment at the current time step.
\end{itemize}
Both accumulated rewards characterize how satisfied users and content providers are on the platform, and number of viable providers reflects how agents help  less established providers.
In all the plots below, we  show the average and the standard error (error bar) of these statistics across test rollouts.

\subsection{When, and Why,  \texttt{EcoAgent} helps both Provider and User Utilities?}
We start with presenting our results under the main environment setup --  \emph{saturated content provider satisfaction}, which defines content provider satisfaction following Assumption \ref{assump:saturared_content provider}. 

\begin{figure}
    \centering
    \includegraphics[width=\linewidth]{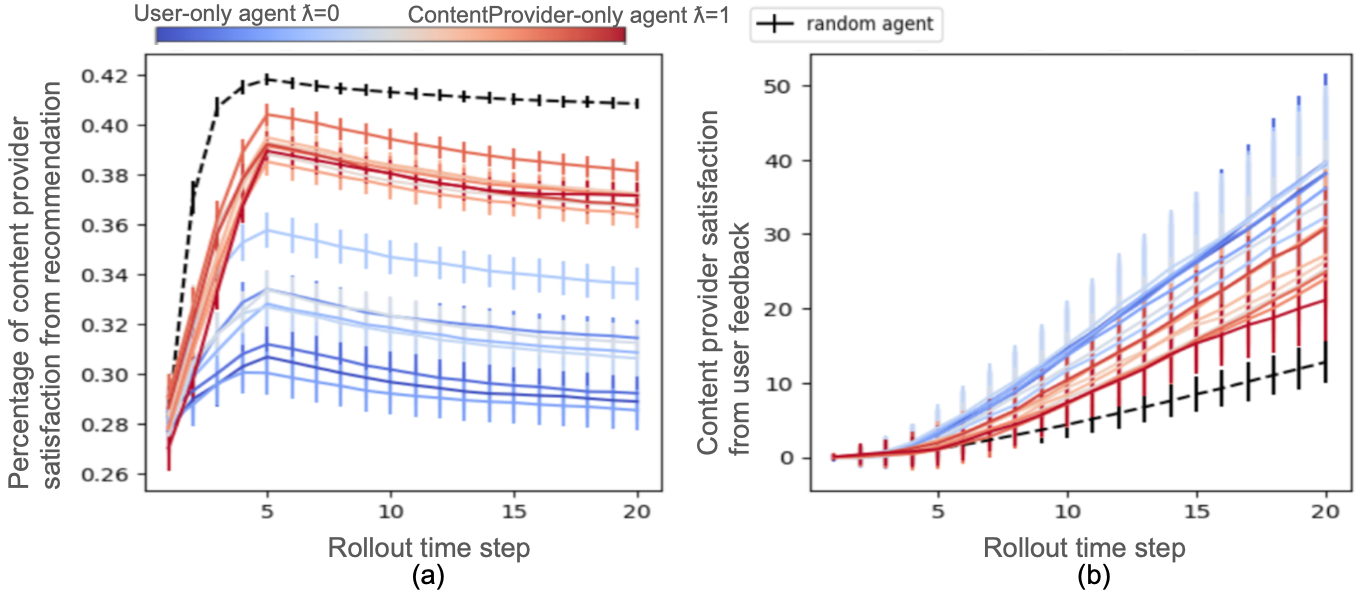}
    \caption{Decomposition of content provider reward with respect to being recommended and user feedback. (a) With a content provider aware \texttt{EcoAgent}, being recommended contributes more to provider utilities as compared to that with a user-only agent. (b) The user-feedback part of the reward is higher for \texttt{EcoAgent}s compared to a random agent, as the latter ignores user preference.}
    \label{fig:submodular_random}
\end{figure}

\begin{figure}
    \centering
    \includegraphics[width=\linewidth]{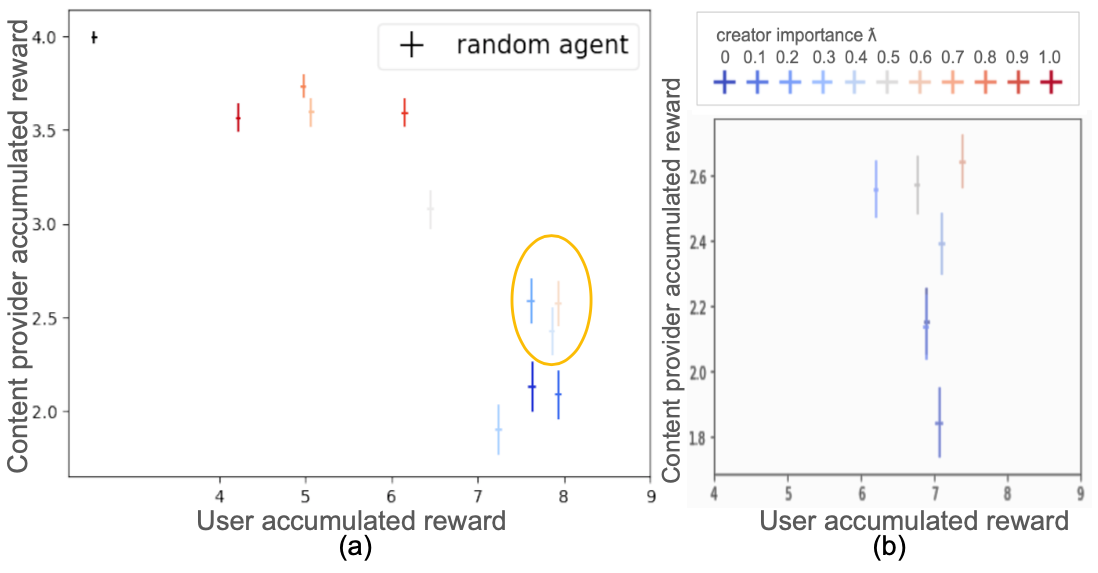}
    \caption{Tradeoff between users and content providers accumulated reward. (a) A properly tuned content provider aware \texttt{EcoAgent} can have a good balance between user utility and content provider utility. (b) In a setup where there exist two identical groups of providers, differing only in initial satisfaction, as expected, there is no tradeoff---we can increase provider utility without sacrificing user utility.}
    \label{fig:submodular_content provider_user} 
\end{figure}
\vskip 2mm
\noindent \textbf{Does \texttt{EcoAgent} increase content provider satisfaction?} 
The first question is if the consideration of content provider utility uplift can indeed lead to happier content providers. We vary the content provider constant $\lambda$ from $0$ (user-only \texttt{EcoAgent}) to $1$ (content provider-only \texttt{EcoAgent}) as shown in Figure \ref{fig:submodular_content provider_help}. It can be seen that content provider accumulated reward (Figure \ref{fig:submodular_content provider_help}a) and number of viable content providers (Figure \ref{fig:submodular_content provider_help}b) are increased with content provider oriented \texttt{EcoAgent}s ($\lambda$ close to $1$) as compared to user-oriented \texttt{EcoAgent} ($\lambda$ close to $0$). Particularly, content provider-only \texttt{EcoAgent} ($\lambda=1$) generates recommendations only to maximize content provider utility and thus has largest content provider satisfaction as well as most viable content providers. 

Recall that content provider reward is decided by number of recommendations and user feedback on the recommended content. 
Figure \ref{fig:submodular_random}a shows the percentage of being recommended contributing to provider reward, and  Figure \ref{fig:submodular_random}b shows the total user feedback contributing to provider reward.
With a random agent, recommendations are more evenly distributed among content providers, which results in a noticeable portion of  provider reward coming from being recommended (Figure \ref{fig:submodular_random}a); larger as  compared to \texttt{EcoAgent}s. However, a random agent does not consider user preference, and thus the user-feedback portion of  provider reward is significantly lower than \texttt{EcoAgent}s as demonstrated in Figure \ref{fig:submodular_random}b.  

\vskip 2mm

\noindent \textbf{Will user satisfaction be sacrificed?} 
The second question comes naturally given that \texttt{EcoAgent} improves content provider satisfaction---then how about users? As the Pareto curve in Figure \ref{fig:submodular_content provider_user}a shows, a heavily content provider aware \texttt{EcoAgent} ($\lambda$ close to $1$), though improving much on content provider accumulated reward, has noticeably smaller user accumulated reward 
as compared to a user-only \texttt{EcoAgent}, which is yet still better than a random agent. However, \texttt{EcoAgent}s in the yellow circle of Figure \ref{fig:submodular_content provider_user}a suggests that, for a properly content provider aware \texttt{EcoAgent} ($0<\lambda\leq 0.6$), we can improve content provider satisfaction without sacrificing user satisfaction much. What is remarkable is that, with certain $\lambda$ parameter choices, not only content provider reward is increased, but also user reward is higher compared to a user-only agent ($\lambda=0$).

\begin{figure}
    \centering
    \includegraphics[width=\linewidth]{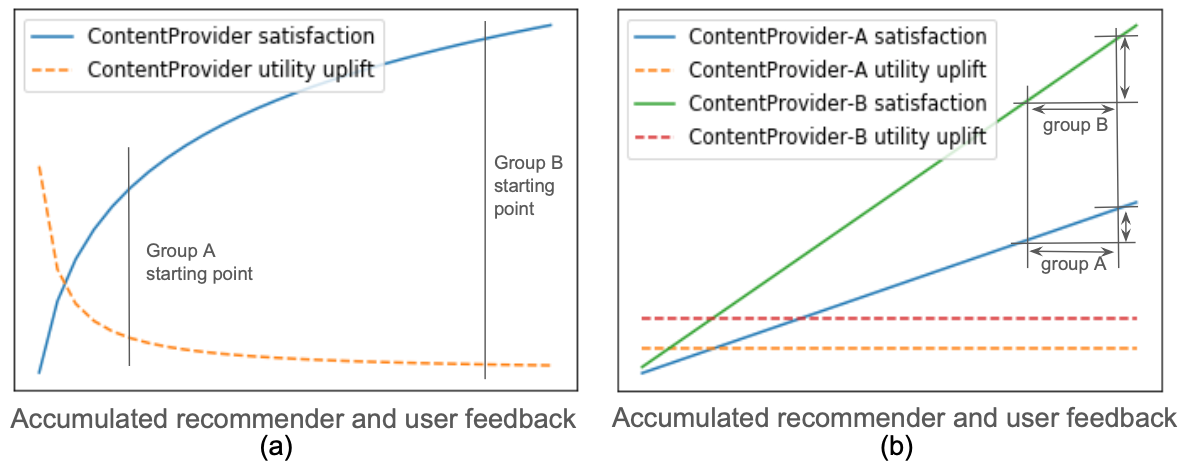}
    \caption{Illustration of environment setups with subgroup content providers. Figure (a) shows an environment of saturated content provider satisfaction, which is initialized with two identical content provider subgroups except that they have different starting points for satisfaction. Figure (b) shows an  environment of linear content provider satisfaction, initialized with two identical content provider subgroups except that they have different rate in increasing satisfaction with incremental recommender and user feedback.}
    \label{fig:subgroup}
\end{figure}

To gain further understanding on the trade-off between user and provider utilities, we consider an additional environment setup where there are two provider groups, who have identical features except for the initial content provider satisfaction---Group A has a smaller starting point, while group B has a larger starting point, as shown in Figure \ref{fig:subgroup}a. Conceptually, group A should be recommended more by \texttt{EcoAgent}, since doing so brings more content provider utility uplift; but this should not impact user satisfaction much, since these two groups are identical except the initial content provider satisfaction. Figure \ref{fig:submodular_content provider_user}b verifies our conjecture: these \texttt{EcoAgent}s with $\lambda\leq 0.6$  increase provider accumulated reward without sacrificing user accumulated reward, that is, a content provider aware \texttt{EcoAgent} with an appropriate provider constant can achieve a good balance between user and content provider satisfactions, verifying the promise of the approach in this setting. 

\begin{figure}
    \centering
    \includegraphics[width=.5\linewidth]{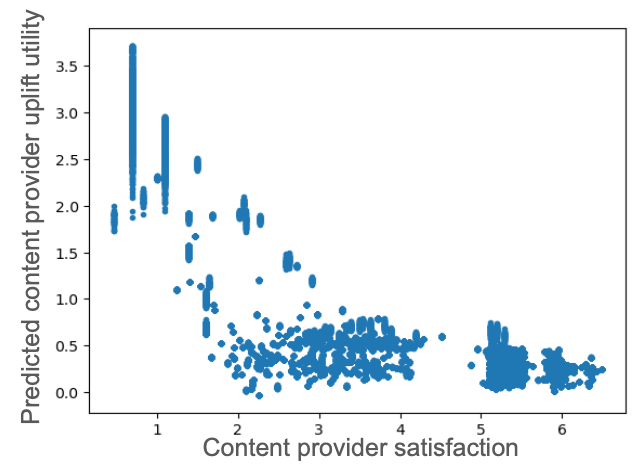}
    \caption{Content provider utility uplift versus content provider satisfaction. In an environment with log (and thus saturated) content provider satisfaction shown in Figure \ref{fig:linear-log-utility}a,  predicted content provider utility uplift by \texttt{EcoAgent}  is inversely proportionally to content provider satisfaction. \texttt{EcoAgent} promotes  less established content providers more to gain more content provider utility uplift, which thus increases  overall content provider utilities. }
    \label{fig:submodular_content provider}
\end{figure}

\vskip 2mm
\noindent \textbf{Why does content provider utility uplift help?} We next examine the rationale behind content provider utility uplift. In an environment with saturated content provider satisfaction, 
the predicted content provider utility uplift by \texttt{EcoAgent} is inversely proportional to content provider satisfaction as shown in Figure \ref{fig:submodular_content provider}. In such cases, less established content providers 
have larger ground truth content provider utility uplift as compared to established content providers (Figure \ref{fig:linear-log-utility}a). By predicting content provider utility uplift,  \texttt{EcoAgent} can identify and promote those less established content providers who need recommendation most, which increases overall content provider utilities and also helps platform have more content providers viable, as we already discussed in Figure \ref{fig:submodular_content provider_help}.

\subsection{When, and why, \texttt{EcoAgent} does \emph{not} lead to a healthier system?}
Until this point, one can clearly see that \texttt{EcoAgent} helps content providers, as long as  content provider satisfaction gets saturated. It raises a natural question on how sensitive \texttt{EcoAgent} is to environmental setup and what happens in the opposite scenarios. 

\begin{figure}
    \centering
    \includegraphics[width=.65\linewidth]{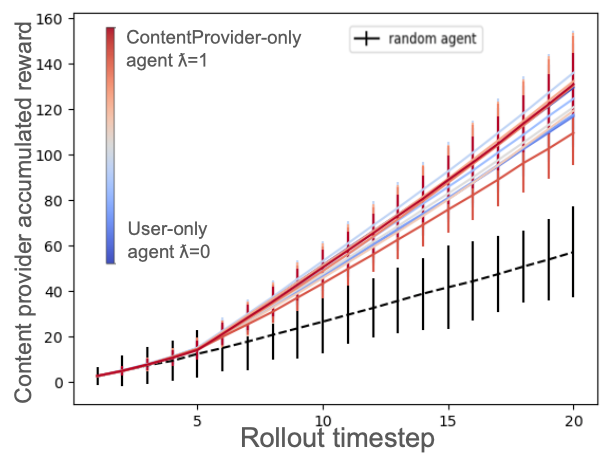}
    \caption{Performance of \texttt{EcoAgent}s when  content provider satisfaction does not get saturated.  In an environment of linear provider-satisfaction shown in Figure \ref{fig:linear-log-utility}b, \texttt{EcoAgent}s with varying $\lambda$ (content provider constant) achieve similar results on provider accumulated reward. This is because there is no difference on provider uplift utilities among different providers, so that \texttt{EcoAgent}s cannot identify less established  providers versus established providers.}
    \label{fig:linear_content provider}
\end{figure}
\vskip 2mm
\noindent\textbf{What if content provider satisfaction does not get saturated?} We  consider an environment with \emph{linear content provider satisfaction}. As opposed to saturated content provider satisfaction function $f$ such as logarithm, here we consider $f$ to be linear, where content provider satisfaction $S^c_t$ and accumulated recommender and user feedback $\sum_{s=1}^t p_t^c$ are exactly in proportion: $S^c_t=f(\sum_{s=1}^t p_s^c) = \eta \sum_{s=1}^t p_s^c$, as demonstrated in Figure \ref{fig:linear-log-utility}b. Content provider utility uplift thus remains the same across different content providers. As expected, \texttt{EcoAgent} would not help here as shown in Figure \ref{fig:linear_content provider}, since there is no guidance for \texttt{EcoAgent} to distinguish less established content providers from established content providers.

\begin{figure}
    \centering
    \includegraphics[width=\linewidth]{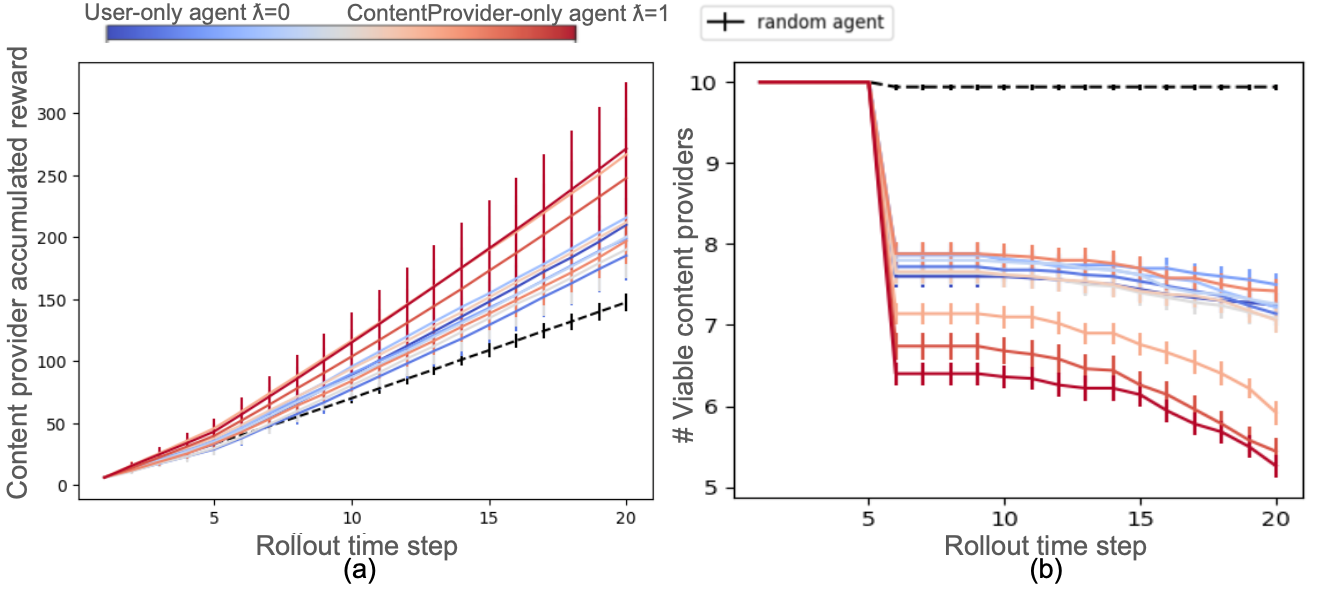}
    \caption{Performance of \texttt{EcoAgent}s for subgroups of content providers with different linear satisfaction functions. 
    Consider an environment with subgroups of content providers who have different rates in increasing satisfaction with respect to accumulated recommender and user feedback, as shown in Figure \ref{fig:subgroup}(b).
    (a) \texttt{EcoAgent} improves content provider accumulated reward as compared to a user-only agent. (b) However, this reduces number of viable content providers on the platform, which results from the \texttt{EcoAgent}'s unfair favor of those content providers who are more sensitive to environment but otherwise have no difference from the others. }
    \label{fig:linear_subgroup}
\end{figure}

\begin{figure}
    \centering
    \includegraphics[width=\linewidth]{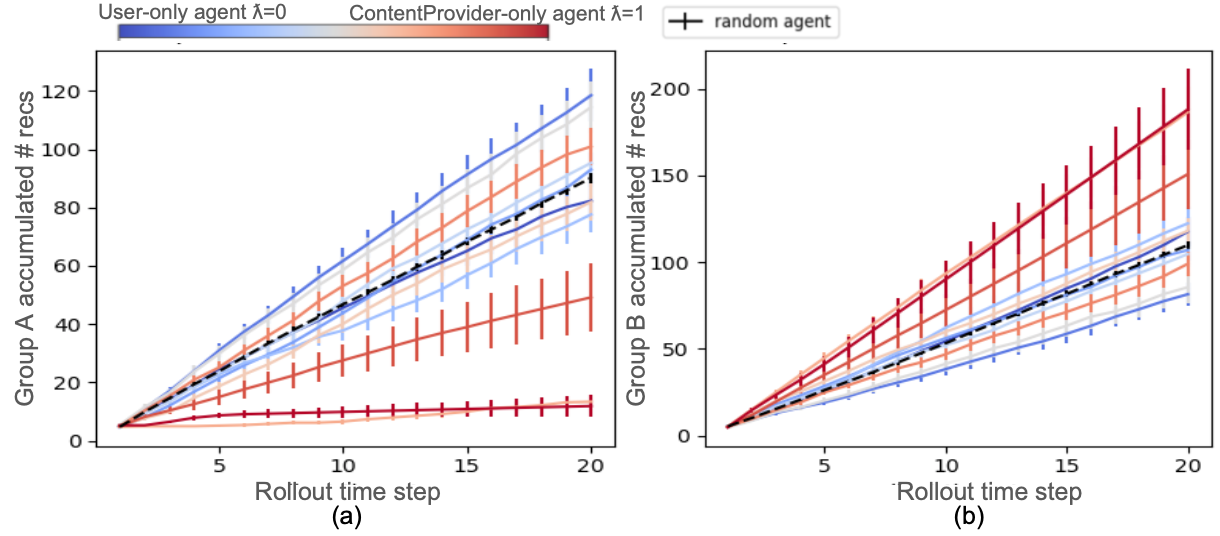}
    \caption{Accumulated recommendations of content provider subgroups A and B of Figure \ref{fig:subgroup}(b): (a) for group A with the slower satisfaction increasing rate, (b) for group B, with the faster satisfaction increasing rate. 
\texttt{EcoAgent}s with higher $\lambda$ (content provider constant) learn to recommend group B more, since doing so results in larger provider utility uplift; this in turn leads to more group B viable providers, but fewer group A viable providers---with a decreased number of total (group A+B) viable providers.
    }
    \label{fig:linear_subgroup_why}
\end{figure}

\vskip 2mm
\noindent \textbf{Is there any setup that content provider aware \texttt{EcoAgent} performs differently to a user-only \texttt{EcoAgent}, under the linear content provider satisfaction setting?} The answer is positive as long as content providers with different satisfaction have different increasing rates with respect to accumulated recommender and user feedback.  This allows EcoAgents to identify and recommend content providers with larger satisfaction increasing rate, which brings more provider utility uplift.
To achieve this, we again  consider two identical  provider groups with linear  provider satisfaction, expect that the slope of each  provider  satisfaction function differs: 
\begin{itemize}
    \item group A content providers: $S^{c_A}_t=\eta_A \sum_{s=1}^t p_t^{c_A}$;
    \item group B content providers: $S^{c_B}_t=\eta_B \sum_{s=1}^t p_t^{c_B}$,
\end{itemize}
where $\eta_B > \eta_A$ such that group B content providers increase their utilities faster than group A content providers, as shown in Figure \ref{fig:subgroup}b. By design, \texttt{EcoAgent} can identify these two groups based on predicted content provider utility uplift---group B should have larger uplift. Figure \ref{fig:linear_subgroup}a shows that content provider aware \texttt{EcoAgent}s increase content provider satisfaction as compared to a user-only \texttt{EcoAgent}. This is because \texttt{EcoAgent}s can gain more content provider utility uplift by promoting group B more (shown in Figure \ref{fig:linear_subgroup_why}), thus improving overall content provider satisfaction. \\\\
\noindent \textbf{However, does \texttt{EcoAgent} really help keep many viable content providers this time?} Surprisingly,  Figure \ref{fig:linear_subgroup}b shows that the number of viable content providers with content provider aware \texttt{EcoAgent}s instead decreases, even though these \texttt{EcoAgent}s have larger content provider accumulated reward as compared to a user-only \texttt{EcoAgent} (Figure \ref{fig:linear_subgroup}a). The problem originates from the incapability of utility uplift  to  capture the difference between less established content providers and established content providers, when the saturated satisfaction is violated.
This leads to our reflection on \texttt{EcoAgent} objective, which uses content provider utility uplift to  optimize overall content provider utilities. Such objective is rational as long as  content provider utility can fully characterize the happiness of each content provider individual, or the platform fairness with respect to content providers. However, \emph{if one's ultimate goal is to maintain a healthy multi-stakeholder platform of more content providers and users, only resorting to maximizing content provider utility can be biased}.  In such cases, \texttt{EcoAgent} objective needs to be adjusted correspondingly, such as including \emph{content provider viability uplift}---the difference of content provider viability probabilities between being recommended versus not. 

\section{Discussion}\label{sec:discussion}
In this paper we study a content provider aware recommender agent that aims at maximizing the combined user and content provider utilities, which we refer to as \texttt{EcoAgent}. We show that maximizing the utility uplift of the provider associated with the chosen content is equivalent to maximizing utilities of all relevant content providers under a mild condition, which merits the scalability of our algorithm. To evaluate \texttt{EcoAgent}, we develop a Gym environment that allows for complex interactions among users, content providers, and agent. We conducted a series of experiments to identify scenarios under which a content provider aware RL recommender can lead to longer term user satisfaction and healthier multi-stakeholder recommender systems. 

A number of important research directions remain open. For example, how to adapt \texttt{EcoAgent} to optimize different metrics, such as number of viable providers or the number of content items. One can modify the lift measurement of utility to alternatives such as modeling the lift of viability probability or content uploading frequencies. 
Another interesting direction is to consider interference among providers, which violates the no-content-provider-externality Assumption \ref{assump:no_content provider_externality}. 

Together, we hope that this study can motivate future research in understanding the impact of recommendation on different entities on the recommendation platforms and building healthy multi-stakeholder recommender systems. 


\bibliographystyle{ACM-Reference-Format}
\bibliography{reference}

\appendix

\end{document}